# BCH-NLP at BioCreative VII Track 3 – medications detection in tweets using transformer networks and multi-task learning

Dongfang Xu[1,2], Shan Chen[1], and Timothy Miller[1,2]
[1]Computational Health Informatics Program, Boston Children's Hospital, Boston, USA
[2]Department of Pediatrics, Harvard Medical School, Boston, USA

*Abstract*— **In this paper, we present our work participating in the BioCreative VII Track 3 - automatic extraction of medication names in tweets, where we implemented a multi-task learning model that is jointly trained on text classification and sequence labelling. Our best system run achieved a strict F1 of 80.4, more than 10 points higher than the average score of all participants. Our analyses show that the ensemble technique, multi-task learning, and data augmentation are all beneficial for medication detection in tweets.**

## I. Introduction

Social media posts have been widely used for monitoring health trends [1–3] and mining opinions [4], largely due to the vast amounts of real-time data and the low barrier to access it. Despite these advantages, social media posts are online-based texts with considerable noise such as jargon, misspellings, abbreviations, emoticons, colloquial language, ungrammatical structures, etc., posing challenges for applying natural language processing (NLP) techniques to mine useful information. In this paper, we present our participation in BioCreative VII Track 3 - automatic extraction of medication names in tweets, one of a series of Social Media Mining for Health Applications (#SMM4H) shared tasks [2,5,6]. This year's task focuses on extracting textual mentions of medications or dietary supplements in tweets, and the dataset used in the shared task exhibits a real distribution of medication mentions in typical Twitter user timelines, containing imbalanced training examples without a predefined list of medications.

We approach the shared task in a multi-tasks setting, where we jointly train a binary classifier and a sequence labelling classifier with a shared transformer network. Our binary classifier takes the representation of the CLS token as input to classify whether a post contains medications, and our sequence labelling classifier takes the representations of all tokens and assigns a label to each one of them. To handle the data imbalance issue, we augment the standard training set from the shared task by including the dataset from the 2018 shared task and extracting extra tweets containing medications from Twitter. Our best system run achieves 80.4 of strict F1, around 10% higher than the mean of all the participants' scores.

## II. Task Description

This shared task aims to find drug products and dietary supplements from tweets. The mentions of drug products come in various forms including trademark names (e.g., NyQuil), generic names (e.g., acetaminophen), and class names (e.g., antibiotics or seizure medication). The mentions of dietary supplements are the names of nutrients added to the diet, which are mainly vitamins. The dataset from the shared task (2021 dataset) consists of all tweets posted by 212 Twitter users during their pregnancy: 88,988 tweets from the training set with 218 tweets mentioning at least one drug, 38,137 tweets from the dev set with 93 tweets mentioning at least one drug, and 54,482 tweets from the test set. This data represents the natural and highly imbalanced distribution of drug mentions on Twitter, with only approximately 0.2% of the tweets mentioning a drug or dietary supplement. One thing to note is that the train/dev/test split covers the posts from all 212 Twitter users, while only 104 users mention drugs in their posts on train + dev splits.

## III. Method

### A. Preprocessing

For preprocessing, we firstly use the open-source tool NLTK TweetTokenizer[1] to tokenize each tweet into tokens. As hashtag is a single token after tokenization, we further split the hashtag into hash sign "#" and its following tag string. To reduce the noise of social media texts, we replace all URLs with token "URL" and all user mentions with the "@USER" placeholder. Lastly, we remove all the special characters that cannot be recognized by the WordPiece tokenizer from the Huggingface transformers library. After the tokenization, We use the BIO tagging schema to generate the label for each token.

### B. Data Augmentation

To mitigate the challenges of imbalanced training examples with no predefined medications, we consider two ways of data augmentation: 1) processing the existing dataset from the SMM4h-2018 shared task [6]; 2) extracting extra tweets mentioning medications from Twitter. For all the augmented tweets, we follow the same preprocessing pipelines.

---

[1]. https://www.nltk.org/api/nltk.tokenize.html

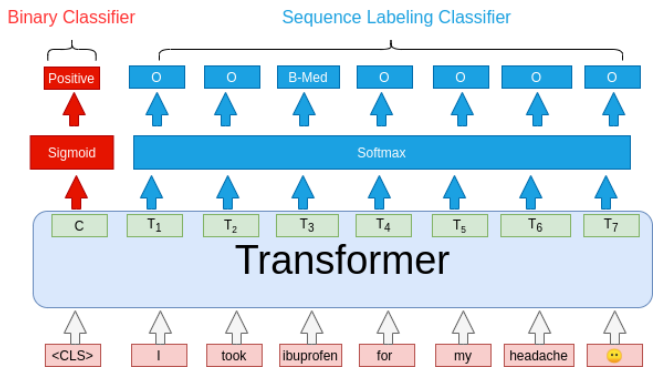

Fig. 1. Our proposed mult-task learning model.

In addition to the 2021 dataset, the task organizers also released the 2018 dataset which was used in a binary classification task aiming to detect whether the tweets contain drug mentions. The 2018 dataset has a balanced distribution with 4975 tweets containing drug mentions (positive tweets) and 4647 tweets without any drug mentions (negative tweets). However, the annotations only contain binary labels and normalized drug names, and such normalized terms may not appear in tweets due to the lexical variants of drug mentions. To find the spans of the drug mentions from positive examples, we conduct both exact and partial string matching techniques to find the offsets for each mention. One thing to note is that most negative examples in the 2018 dataset contain ambiguous mentions that can be confused with medication [6].

We also used the Twitter API[2] to extract additional tweets by keyword searching. We first collect the top 200 frequent drug mentions from the 2021 and 2018 datasets as the keywords. We then extract 100 tweets for each of these terms from Twitter. As we found the majority of tweets in the 2021 datasets have fewer than 128 characters, we remove the tweets that are longer than 128 characters. We further use the self-training framework to select the examples with the most confident drug span predictions. Specifically, we fine-tune the BioRedditBERT-base-uncased [7] model described in the following section on the 2018 dataset and the 2021 dataset, and make predictions on the augmented tweets. We then select around 77,000 tweets whose predictions only include one drug mention and their span prediction scores are higher than 0.9. To increase the coverage of different drug mentions while also maintaining a similar distribution of drug uses, we manually annotate the use categories for the top 200 drugs, including pain, heartburn, birth control, cold, etc., and collect 326 similar drugs in total belonging to these use categories from the WebMD[3]. And for each one of the 326 drugs, we randomly select 4 tweets from the 77,000 tweets mentioning drugs with the same use categories and replace their original drug mention with it. After removing the duplicate tweets, we eventually obtained 1194 tweets.

*C. Model Description*

Inspired by the baseline system [8], where it first applies a text classifier to prefilter the tweets without mentioning drugs and then uses a sequence labelling classifier to extract the spans of drug mentions from the remaining tweets, we implement a multi-task learning model to tackle the text classification and sequence labelling jointly (see figure 1). We first feed the tweet as input to the transformer network to generate a representation for each subword token; the sigmoid classifier takes the representation of the CLS token to output a binary label, and the softmax classifier takes the representations of all other subword tokens and outputs one BIO tag for each input. The final loss of our model is the sum of losses from two classifiers. In contrast to the model just trained on the sequence labelling task, we hope the jointly training model can better leverage the context information for the sequence labelling classifier.

*D. Ensemble System*

In an attempt to improve performance over individual models, we apply an ensemble technique to combine the results of multiple different models. We select 8 transformer models pre-trained with different settings as our initialization points: BERT-base-cased, BERT-base-uncased, BERT-large-cased, BERT-large-uncased, BioBERT-base-cased, BioRedditBERT-base-uncased, BioClinicalBERT-base-cased, PubMedBERT-base-uncased. Among these models, BERT- [9] models are pre-trained on the general domain texts; BioBERT- [10] model is initialized with weights from BERT- and further pre-trained on PubMed abstracts and PMC full-text articles; BioClinicalBERT- [11] is initialized with weights from BioBERT- and further pre-trained on MIMIC III clinical notes; BioRedditBERT- [7] is initialized with weights from BioBERT- and further pre-trained on health-related Reddit posts; PubMedBERT- [12] is pretrained from scratch using abstracts from PubMed and full-text articles from PubMedCentral. -base- and -large- are models with 12 or 24 transformer layers, respectively. -cased- models are sensitive to the case information from inputs, while -uncased models are not.

After fine-tuning the models on tweets, we aggregate the outputs from the above eight models (i.e., the spans of drug mentions) and select predictions that have been agreed upon by multiple models, where the exact number is a parameter we tune on development data.

*E. Experiments*

Unless specifically noted otherwise, we keep the default hyper-parameters as in huggingface's PyTorch [4] implementation (transformers-4.4.1) across all experiments. We fine-tune each one of the 8 models for 10 epochs with learning rate of 3e-5, batch size of 64, maximal sequence length of 128, and the checkpoints that achieve the best performances on the dev set are selected as one of the ensemble models. We fine-tune the models on three different data combinations: 1) 2021 dataset + positive examples from the 2018 dataset + extra extracted tweets (2021+2018_positive+extra); 2) 2021 dataset + 2018 dataset + extra extracted tweets (2021+2018+extra); 3) 2021 dataset + 2018 dataset (2021+2018). We also tried fine-tuning models on the 2021 dataset with upsampling or downsampling techniques, but these resulted in worse performances, and are sensitive to different hyperparameters.

---

[2] https://developer.twitter.com/en/docs
[3] https://www.webmd.com/drugs/2/index
[4] https://github.com/huggingface/transformers

TABLE I. PERFORMANCES OF OUR SYSTEM RUNS ON THE TEST SET OF THE SHARED TASK

|  | Strict | | | overlap | | |
|---|---|---|---|---|---|---|
|  | P | R | F1 | P | R | F1 |
| Baseline system [8] | 89.0 | 66.0 | 75.8 | 90.8 | 67.3 | 77.3 |
| Avg. score of all participants | 75.4 | 65.8 | 69.6 | 81.1 | 70.9 | 74.9 |
| System run 1 | 83.0 | 76.2 | 79.4 | 85.2 | 79.4 | 81.6 |
| System run 2 | 79.9 | 81.0 | 80.4 | 81.9 | 83.0 | 82.4 |
| System run 3 | 81.0 | 78.2 | 79.6 | 84.5 | 81.6 | 83.0 |

During the evaluation period, we submitted three system runs: 1) outputs from an ensemble of models fine-tuned on 2021+2018+extra and agreed by more than six models; 2) outputs the same setting as 1) but agreed by more than five models; 3) outputs from an ensemble of models fine-tuned on 2021+2018 and agreed by more than five mode.

## IV. RESULTS AND DISCUSSION

Table I shows that all our three system runs outperform the baseline system, mostly because our system runs have better recall and balanced precision-recall tradeoff. Our system run 2 achieves the best strict F1 of 80.4 among all our submissions, 4.6 points higher than the baseline system, and 10.6 points higher than the mean of all participants. Comparing system run 2 against 1 and 3, we find that decreasing the threshold of how many models agree on the predictions (6 vs. 5), and training on the extra tweets, had better recall without harming precision too much.

To understand the effects of model initializations, data augmentations, and ensemble techniques, we present the performances of different models or ensemble systems in table II. We first analyze the effects of different data combinations: firstly, 7 out of 8 models trained on 2021+2018+extra have better strict F1 scores than models trained on 2021+2018_positive+extra or 2021+2018; secondly, comparing models trained on 2021+2018+extra and 2021+2018_positive+extra, we find the ambiguous negative examples from the 2018 datasets improve the precision for 6 out of 8 models; lastly, comparing models trained on 2021+2018+extra and 2021+2018, we find our extra extracted tweets improve the recall for 7 out of 8 models. In terms of the model initializations, we find that -uncased models have better average ranks than -cased models, which indicates that ignoring the case information is helpful for drug detection in tweets, especially on models pre-trained on the general domain. We also find that models further pre-trained on biomedical texts are helpful for the task, most likely because the biomedical terms are good signs for detecting drug terms.

Lastly, our ensemble systems have better strict F1 than each of their individual systems. Compared with each individual model, the performance gaps of the ensemble systems fine-tuned on different data combinations are much smaller, indicating that the ensemble systems are less sensitive to the different data combinations.

TABLE II. THE STRICT PERFORMANCES OF EIGHT DIFFERENT MODELS AND THEIR ENSEMBLE SYSTEMS WHEN FINE-TUNED ON THREE DIFFERENT DATA COMBINATIONS AND EVALUATED ON THE DEV SET. THE AVG. RANK COLUMN SHOWS THE AVERAGE RANK OF EACH FINE-TUNED MODEL ACROSS MULTIPLE DATA COMBINATIONS. MODELS WITH THE BEST STRICT F1 ARE BOLDED. THE ENSEMBLE-5 AND ENSEMBLE-6 ROWS ARE THE ENSEMBLE SYSTEM WITH 5 AND 6 INDIVIDUAL MODELS AGREEING ON THE FINAL PREDICTIONS.

|  | 2021+2018+extra | | | 2021+2018_positive+extra | | | 2021+2018 | | | AVG. Rank |
|---|---|---|---|---|---|---|---|---|---|---|
|  | P | R | F1 | P | R | F1 | P | R | F1 |  |
| BERT-base-cased | 76.6 | 90.5 | 83.0 | 71.2 | 89.5 | 79.3 | 74.6 | 81.0 | 77.6 | 7.3 |
| BERT-base-uncased | 85.2 | 87.6 | 86.4 | 75.4 | 90.5 | 82.3 | 77.9 | 83.8 | 80.7 | 3.7 |
| BERT-large-cased | 81.2 | 86.7 | 83.9 | 70.1 | 89.5 | 78.7 | 76.9 | 85.7 | 81.1 | 6 |
| BERT-large-uncased | 80.2 | 92.4 | 85.8 | 82.3 | 88.6 | **85.3** | 83.2 | 84.8 | 84.0 | 2 |
| BioBERT-base-cased | 79.3 | 87.6 | 83.3 | 73.6 | 90.5 | 81.2 | 79.5 | 84.8 | 82.0 | 4.6 |
| BioRedditBERT-base-uncased | 88.5 | 87.6 | **88.0** | 74.0 | 89.5 | 81.0 | 83.6 | 87.6 | **85.6** | 2.6 |
| BioClinicalBERT-base-cased | 79.8 | 90.5 | 84.8 | 73.6 | 90.5 | 81.2 | 78.4 | 82.9 | 80.6 | 5 |
| PubMedBERT-base-uncased | 70.8 | 92.4 | 80.2 | 74.6 | 92.4 | 82.6 | 80.9 | 84.8 | 82.8 | 4.3 |
| Ensemble-5 | 90.6 | 89.2 | 89.8 | 87.0 | 89.5 | 88.3 | 92.6 | 85.7 | 89.1 | - |
| Ensemble-6 | 91.2 | 88.6 | 89.9 | 85.5 | 89.5 | 87.4 | 94.6 | 82.9 | 88.3 | - |

TABLE III. PERFORMANCES OF TWO BEST INDIVIDUAL MODELS WHEN FINE-TUNED ON 2021+2018+EXTRA WITH MULTI-TASK OBJECTIVES (TEXT CLASSIFICATION + SEQUENCE LABELLING) AND SINGLE-TASK OBJECTIVES (SEQUENCE LABELLING).

| Training objective | Text classification + Sequence labelling | | | Sequence labelling | | |
|---|---|---|---|---|---|---|
| | P | R | F1 | P | R | F1 |
| BioRedditBERT-base-uncased | 88.5 | 87.6 | **88.0** | 78.4 | 90.9 | 84.2 |
| BERT-large-uncased | 80.2 | 92.4 | **85.8** | 80.2 | 89.5 | 84.6 |

We also conduct an ablation analysis to understand the effect of multi-tasks learning. We select two best individual models and train them with/without multi-task learning. Table III shows that jointly training the text classifier and sequence labelling classifier is beneficial for drug mentions detection.

## V. CONCLUSION

In this paper, we present our work participating in the BioCreative VII Track 3 - automatic extraction of medication names in tweets, where we implement a multi-task learning model that is jointly trained on text classification and sequence labelling. Our best system run achieved a strict F1 of 80.4, more than 10 points higher than the average score of all participants. Our analyses show that the ensemble technique, multi-task learning, and data augmentation are beneficial for medication detection in tweets.